\documentclass{article}

\usepackage{arxiv}

\usepackage{times}
\usepackage{epsfig}
\usepackage{graphicx}
\usepackage{amsmath}
\usepackage{amssymb}
\usepackage{multirow}
\usepackage{fancyref}
\usepackage{adjustbox}
\usepackage{booktabs}
\usepackage{subcaption}
\usepackage[ruled]{algorithm2e}
\newcommand{\R}{\mathbb{R}}

\usepackage{cite}
\usepackage{amsmath,amssymb,amsfonts}
\usepackage{algorithmic}
\usepackage{graphicx}
\usepackage{textcomp}
\usepackage{xcolor}
\def\BibTeX{{\rm B\kern-.05em{\sc i\kern-.025em b}\kern-.08em
    T\kern-.1667em\lower.7ex\hbox{E}\kern-.125emX}}

\begin{document}

\title{The Taboo Trap: Behavioural Detection of Adversarial Samples}

\author{
Ilia Shumailov\thanks{Equal contribution} \\
  Computer Laboratory\\
  University of Cambridge\\
  \texttt{ilia.shumailov@cl.cam.ac.uk} \\
  %% examples of more authors
   \And
  Yiren Zhao* \\
  Computer Laboratory\\
  University of Cambridge\\
  \texttt{yiren.zhao@cl.cam.ac.uk} \\
  \And
  Robert Mullins \\
  Computer Laboratory\\
  University of Cambridge\\
  \texttt{robert.mullins@cl.cam.ac.uk} \\
  \And
  Ross Anderson \\
  Computer Laboratory\\
  University of Cambridge\\
  \texttt{ross.anderson@cl.cam.ac.uk} \\
}

%Ilia Shumailov∗, Yiren Zhao∗, Robert Mullins, Ross Anderson} 

\maketitle

\begin{abstract}
Deep Neural Networks (DNNs) have become a powerful tool
for a wide range of problems.
Yet recent work has found an increasing variety of adversarial samples
that can fool them.
Most existing detection mechanisms against adversarial attacks
impose significant costs,
either by using additional classifiers to spot adversarial samples,
or by requiring the DNN to be restructured.
In this paper, we introduce a novel defence.
We train our DNN so that, as long as it is working
as intended on the kind of inputs we expect,
its behavior is constrained, in that some set of behaviors are taboo.
If it is exposed to adversarial samples,
they will often cause a taboo behavior, which we can detect.
% As an analogy,
% we can imagine that we are teaching our robot good manners;
% if it's ever rude, we know it's come under some bad influence.
% This defence mechanism is very simple and has almost zero computation overhead at runtime
% -- making it particularly suitable for defending cheap attacks at scale.
% Just as humans' choice of language can convey a lot of information about
% location, affiliation, class and much else that can be opaque to
% outsiders but that enables members of the same group to recognise each other,
Taboos can be both subtle and diverse,
so their choice can encode and hide information.
% We can use this to make adversarial attacks much harder.
It is a well-established design principle that
the security of a system should not depend on the obscurity of its design,
but on some variable (the key) which can differ between implementations and be
changed as necessary.
We discuss how taboos can be used to equip a classifier
with just such a key,
and how to tune the keying mechanism to adversaries of various capabilities.
We evaluate the performance of a prototype against a wide range of attacks and
show how our simple defense can defend against cheap attacks at scale
with zero run-time computation overhead, making it a suitable
defense method for IoT devices.
%In addition, the proposed method can be combined with existing defense
%mechanisms to enhance security, making it a ready tool to be added to the defender's armoury.
\end{abstract}

\section{Introduction}

Deep Neural Networks (DNNs) are being built into ever more systems.
Applications such as vision \cite{krizhevsky2012imagenet},
language processing \cite{collobert2008unified}
and fraud detection \cite{patidar2011credit} are now using them.
Some of these applications are starting to support safety-critical missions.
Unfortunately,
recent research has discovered that surprisingly small changes to inputs,
such as images and sounds, can cause DNNs to misclassify.
In the new field of adversarial machine learning,
attackers craft small perturbations that are not perceptible by humans
but are effective at tricking DNNs.
Such adversarial samples can pose a serious threat to
systems that use machine learning techniques:
examples range from tricking a voice or face recognition system
to break into stolen smart phones \cite{Carlini2016},
to changing road signs and street scenes
so that human drivers will interpret them one way
while autonomous cars or drones read them
differently \cite{eykholt2018robust}.

Current adversarial sample detection
has two main approaches.
The first is to train additional classifiers
to distinguish between clean
and adversarial samples \cite{lu2017safetynet, DBLP:journals/corr/abs-1711-08478}.
The second is to restructure or augment the
original network topology \cite{DBLP:journals/corr/GrosseMP0M17}.
These mechanisms detect a range of attacks,
but are not available off-the-shelf and are not infallible.
Both introduce extra design complexity and computational cost
-- which can be particularly difficult for low-cost or energy-limited IoT devices.
Low-cost attacks can be surprisingly portable \cite{zhao2018compress},
so expensive defense mechanisms would impose additional costs on a wide range of products.

In this paper, we introduce a novel method of detecting adversarial samples,
named the {\em Taboo Trap}.
As before,
we detect them by observing the classifiers' behaviour.
However, rather than learning the expected behaviour of a classifier we wish to protect, and then trying to harden it against adversarial samples,
we restrict its behaviour during training and detect the unexpected reactions that such samples cause.
% This work is in line with the observation made by Zhao et al. that
% adversarial attacks are less transferable
% with less numerical ranges available on activations \cite{zhao2018compress}.

% An intuitive analogy is that we educate our children
% to have good manners, and in particular to not use certain words that are taboo in polite company.
% If the kids learn taboo words in the playground, these bad influences are immediately obvious to parents.
% An intuitive explanation is that when an activation representation
% is limited in magnitude, the adversary will struggle to over-drive
% activations to cause a misclassification.
% An intuitive analogy is that we train our robots to behave well,
% and if one of them is suddenly starts being rude to us,
% we know it has fallen under some bad influence --
% as when we educate children to be polite,
% and yet they have been influenced by inappropriate words in the playground
% and use them at the dinner table. By `inappropriate' we do not just mean
% anatomical swearwords or discriminatory epithets;
% everyday vocabulary choice coveys a lot of information about
% social group membership, political views and so on~\cite{Ross54}.
% (The analogy is slightly stretched,
% as in a DNN we can measure not just the output
% but also the activations of intermediate layers;
% it is as if we can detect a subverted child not just
% when they speak an inappropriate word, but even when they think one.)

To set a Taboo Trap, we first profile the activation values and choose a transform function to apply to them so they can be limited to specific ranges. We then re-train the network to get one with the restricted activations we chose.
Finally, we use a detection function that rings the alarm if any activation
breaks a taboo by falling outside the expected ranges.
In a simple implementation,
the transform function can be simply clipping activation values,
which adds almost zero computation overhead, and is effective
at rejecting simple attacks.

In this paper we make the following contributions:
\begin{itemize}
    \item We propose a conceptually different defense strategy, namely restricting DNN behaviour during training so as to detect adversarial inputs later;
    \item We evaluate our proposed detection mechanism using a range of adversarial attacks and DNNs, demonstrating that it has a low run-time overhead;
    % and prevents attacks developed on one network from working on another that was trained with a different key.
    \item We show that the changes made to the training process do not impair the convergence and performance of the tested networks.
\end{itemize}

\section{Related Work}
\label{sec:related_work}
Szegedy et al. \cite{Szegedy2013}
first pointed out that DNNs are vulnerable to adversarial samples and that,
in many cases, humans cannot distinguish between clean and adversarial samples.
That paper sparked off an arms race between adversarial attack and defense.
Defense mechanisms come in a wide variety of flavours.
The earliest defense, {\em adversarial training},
was proposed by Szegedy et al. themselves;
they noticed that, when they fed adversarial samples into the training process,
the classifier became more robust \cite{Szegedy2013}.
Szegedy et al. found however that adversarial samples often transfer between models,
enabling {\em black-box attacks}
where a sample trained on one model is used to attack another.
Variations on this theme were then explored,
Tramer et al. found simple attacks that
defeat the basic, single-step, version of adversarial training.
One type involves a random perturbation to escape local minima,
followed by a linearisation step ~\cite{tramer2017ensemble}.
They proposed {\em ensemble adversarial training} in which a model is trained
against adversarial samples generated on a variety of networks
pre-trained from the same data.
They also described {\em gradient masking} as a defense mechanism,
noting that gradient-based attacks need to measure a gradient and making
this hard to do accurately stops the attacks from working~\cite{tramer2017ensemble}.
% Those defenses, however, were soon shown to be vulnerable to black-box attacks.

The alternative defense approach is adversarial sample detection.
Conceptually, such mechanisms are like intrusion detection systems;
they provide situational awareness, which in turn can enable a flexible defense.
Metzen et al. were among the first to study adversarial sample detection~\cite{metzen2017detecting}.
They augmented the original classifier with an auxiliary network
which they trained to classify the inputs as either clean or adversarial,
and concluded that it could detect adversarial samples.
Grosse et al. proposed two other approaches~\cite{DBLP:journals/corr/GrosseMP0M17}.
First, they found feasible to use statistical test to distinguish adversary
from legitimate sample distribution data distribution;
but such test was compute demanding in practice.
Second, they found that instrumenting the network with an additional class
designed for adversarial sample detection is able to work more efficiently.
Meng and Chen focused on detecting adversarial samples using
an additional classifier \cite{Meng:2017:MTD:3133956.3134057} and designed
the MagNet system that consists of a detector
network and a reformer network.
% MagNet learn to distinguish between normal and adversarial inputs
% by measuring their distance from an expected manifold.
% The reformer network uses auto-encoders to move adversarial samples towards the manifold of legitimate ones, so as to stop them fooling the detectors.
% Although MagNet defeats black-box attacks, it was found to be vulnerable to white-box attacks (where the attacker knows the model's parameters); but its creators claim that using multiple different autoencoders can defeat white-box (or at least grey-box) attacks since the defense behaviour becomes too unpredictable. This appears to be the first paper to invoke the concept of cryptographic diversity as a defence.
Lu et al. presented SafetyNet -- another way of instrumenting a classifier
with an adversarial sample detector \cite{lu2017safetynet}.
It replaces the last layers of the classifier with a quantised ReLU activation
function and uses an RBF-based SVM to classify the activation patterns.
% They found that this allows them to detect adversarial samples reliably and note that there might exist even better objective functions.
Li et al. invented another detection method based on observation of the
last-layer outputs of convolutional neural networks \cite{DBLP:journals/corr/LiL16e}.
They built a cascade classifier to detect unexpected behaviours in this last layer.

% There has also been research on using Bayesian methods to detect adversarial samples by estimating the uncertainty of each image input~\cite{feinman2017detecting}
% or applying transformations to inputs to make detection easier~\cite{AAAI1817408}.

All of the above defences, however, have at least one of the following shortcomings.
First, some require additional classifiers \cite{DBLP:journals/corr/LiL16e,metzen2017detecting},
which means extra computation at runtime.
Second, some require restructuring the original neural network
\cite{ lu2017safetynet, Meng:2017:MTD:3133956.3134057}, which means extra design complexity and possible degradation of functionality.
Our goal is to build a detection mechanism for IoT devices that requires
no additional classifiers,
does not impair the network's performance in the absence of adversarial samples,
and can detect them dependably with almost zero computational overhead.
%The proposed defense is then effective and efficient in preventing cheap attacks at scale.

\section{Methodology}
\label{sec:methodology}

% \subsection{Adversarial Attacks}
%A number of researchers have discovered ways to craft imperceptible perturbations on input samples to create adversarial images that cause DNNs to misclassify~\cite{Szegedy2013, goodfellow2014explaining}.
\subsection{The Taboo Trap}

% \begin{figure}[!ht]
%     \centering
%     \includegraphics[width=\linewidth]{images/jiaozi_flow.pdf}
%     \caption{A 3-stage detection pipeline.}
%     \label{fig:attack_pipeline}
% \end{figure}
% \begin{figure*}[!ht]
% \centering
% \begin{subfigure}{.45\textwidth}
%   \centering
%   \includegraphics[width=\linewidth]{images/resnet_percentiles_clean.png}
%   \caption{Clean data}
%   \label{fig:resnet_percentiles_clean}
% \end{subfigure}%
% \begin{subfigure}{.45\textwidth}
%   \centering
%   \includegraphics[width=\linewidth]{images/resnet_percentiles_fgsm.png}
%   \caption{FGSM ($\epsilon=0.4$) data}
%   \label{fig:resnet_percentiles_adversarial}
% \end{subfigure}
% \caption{Predictive accuracy with different percentiles used with Max detector of ResNet18 on CIFAR10 dataset.}
% \label{fig:resnet_both}
% \end{figure*}

The intuition behind the Taboo Trap is simple.
We train the DNN to have a restricted set of behaviours on activations that
are hidden from the attackers during training, and report any behaviour that later violates these restrictions as an adversarial sample.
We first profile activation values for all samples ($N$) on the training dataset
across different layers ($L$), $A \subset \R^{N  \times L \times X \times Y \times C}$.
Each $A_{l, n}$ is a three-dimensional tensor which is a collection of feature maps,
$A_{n, l} \subset \R^{X \times Y \times C}$, where $X$, $Y$ and $C$ are the feature map's width, height and number of channels respectively.
To set a Taboo Trap, we need to define a transform function $f_t$ on all activations.
During the training phase, we combine the output of the transform function, considering only a batch of data ($B$):
\begin{equation}
    L = L_{SGD} + \lambda \sum_{n=1}^{B} f_t(A_{n})
\end{equation}

$\lambda$ is a hyper-parameter and $L_{SGD}$ is the loss from Stochastic Gradient Descent (SGD).
In the third stage, the output of $f_t(A_n)$ translates to the detection result:% notice now we only consider activations for each input sample ($n$):
\begin{equation}
    \text{Detected} =
    \begin{cases}
    \text{True},  & \text{if } f_t(A_n) \geq 0\\
    \text{False},  & \text{otherwise}
\end{cases}\end{equation}

% \subsection{Using $n$th Percentile as Transform Function}

To give an example,
we study one of the many possible transform functions --
the maximum percentile function.
We will return to the discussion of what makes a good transform function,
and how to diversify them, in \Fref{sec:discussion}.

Consider the activation values,
for each layer $l$ and sample $n$,
we have $A_{n, l} \subset \R^{X \times Y \times C}$,
where $X$, $Y$ and $C$ are the feature map's width, height and number of channels respectively.
We use $\max (A) \subset \R^{N \times L}$
to represent the profiled maximum activations of
$N$ samples on $L$ layers; effectively, $\max (A_{n, l})$
is a single value that is the maximum of $A_{n, l}$.
% For each layer $l$, the profiled maximum activations of
% all samples on the training dataset are represented as $maxA_{[1:N], l} \subset  \R^{N}$.
The profiled maximum activation $\max(A_{[1:N], l})$
is a vector of length $N$, where $N$ is the number of samples.
We first calculate a threshold value for a particular layer:

\begin{equation}
    \alpha_l = g(\max (A_{[1:N], l}))
\end{equation}

We design a function
$\alpha_l = g(\max (A_{[1:N], l})) = \textsf{Percentile}_n(\max (A_{[1:N], l}))$,
where $\textsf{Percentile}_n$ computes the $n$th percentile of the function input.
We then perform fine-tuning on the DNN with a regularizer that penalizes activations larger than the profiled thresholds.
We then design the transform function ($f_t$) in the following form:

\begin{equation}
    \begin{aligned}
    f_t(A_n) & = \\
        &\lambda  \sum_{l=0}^{L-1} \sum_{x=0}^{X_l-1} \sum_{y=0}^{Y_l-1} \sum_{c=0}^{C_l-1}f_p(A_{n, l, x, y, c}, \alpha_{l})
    \end{aligned}
\end{equation}

We define $f_p$ to be:
\begin{equation}
    f_p(a, b) =
    \begin{cases}
    1,  & \text{if } a \geq b\\
    0,  & \text{otherwise}
\end{cases}
\end{equation}

% For the mean-std approach, we use a different regularizer to gradually penalize large activations.
% We design three levels of regularization based on mean and standard deviations (std), for values larger than $mean+3std$,
% it receives a larger penalization and thus helps the DNN to control the range of activations.
% For each layer, we generate the three levels $(\alpha_{1,l}, \alpha_{2,l}, \alpha_{3,l})$:
% \begin{equation}
%     \begin{aligned}
%         & \alpha_{1,l} = f_{\mu}(maxA_{l, [1:N])}) + f_{\sigma}(maxA_{l, [1:N])}) \\
%         & \alpha_{2,l} = f_{\mu}(maxA_{l, [1:N])}) + 2f_{\sigma}(maxA_{l, [1:N])}) \\
%         & \alpha_{3,l} = f_{\mu}(maxA_{l, [1:N])}) + 3f_{\sigma}(maxA_{l, [1:N])}) \\
%         % L = &\\
%         %     & \lambda (RSum(A, \alpha_1) + RSum(A, \alpha_2) + RSum(A, \alpha_3))
%     \end{aligned}
% \end{equation}

% The generated three levels can provide us a loss function, notice, we use $\alpha_{1}$ to represent a vector with length $L$.
% \begin{equation}
%     \begin{aligned}
%         Loss = & \lambda (RSum(A, \alpha_1) + \\
%             & RSum(A, \alpha_2) + RSum(A, \alpha_3))
%     \end{aligned}
% \end{equation}
After finding the number of activations
in each layer that are over-excited, we try to minimize their values using SGD.
The regularizer provides an extra loss
to the loss function of the neural network
and $\lambda$ is a hyperparameter used to control
how hard we are penalizing the neural network for having large activation values.
We call $\lambda$ the alarm rate.

Finally, we use a detection function when we deploy the model.
For any given input $n$, we check the activation values that the neural network produces at every layer.
If any of them is greater than the threshold value for that particular layer, we recognize that particular input as adversarial.

% In fact, this behaviour level detection can work with any arbitrary functions and we discuss the different functions in later sections.

\subsection{Training Strategy}

\label{sec:from_scratch_vs_from_pretrained}
% When training with a regularizer, there are two methodologies to consider.
% One is simply to train the network again from scratch,
% and the other one is using the existing pre-trained model as a starting point for fine-tuning.
% We find that there is a big difference in detection rates between the two.
% \Fref{fig:lenet_finetuning} and \Fref{fig:lenet_from_scratch} show the performance of LeNet5 when trained with the same hyperparameters, fine-tuned and from scratch respectively.
% It can be seen that the detection rates are significantly larger with small FGSM $\epsilon$ values.
% When training the network with a large regularizer,
% the weights have to optimize both the regularization loss and the classification loss.
% For fine-tuning, the former dominates the cost function
% thus the model struggles to balance between the two losses
% and makes convergence challenging.
% In contrast, when training from scratch,
% the regulariser loss is so large that it provides heavy restrictions on the initialized weights.
% In summary, we provide the training methodology of the taboo trap in Algorithm~\ref{alg:training}.
When training with Taboo Trap, we find that it is necessary
to gain some model diversity when re-training the models.
Fine-tuning from a pretrained model can lead to a suboptimal outcome when balancing the cost from the taboo regularization term against the model cost.
We summarize our training strategy in Algorithm~\ref{alg:training},
which effectively is about balancing the training hyperparameters.

\begin{algorithm}[!h]
 \SetAlgoLined
 \KwData{$\lambda$ = Alarm Rate, $\epsilon$ = Learning Rate, $l$ = loss}
 \KwResult{WTC instrumented network with comparable performance}
 Pick initial values for Alarm Rate and Learning Rate\;
 \While{Detection Rate on test data $> 0$}{
  Retrain with $\lambda$ and $\epsilon$ and collect loss $l$ for a number of epochs.\\
  \If{$l$ is not decreasing}{
    Increase $\lambda$\\ Decrease $\epsilon$
   }
 }
 \caption{Taboo Trap instrumented training process}
 \label{alg:training}
\end{algorithm}

\begin{table}[!ht]
\centering
\adjustbox{max width=\textwidth}{
\begin{tabular}{@{}l|l|lll|lll|lll@{}}
\toprule
\multicolumn{1}{c}{\textbf{}} & \multicolumn{1}{c}{\textbf{}} & \multicolumn{3}{c}{\textbf{LeNet5 -- MNIST}} & \multicolumn{3}{c}{\textbf{M-CifarNet -- CIFAR10}} & \multicolumn{3}{c}{\textbf{ResNet18 -- CIFAR10}}  \\ \midrule
& $~~~~~~\theta$  & A              & D              & AD             & A                  & D                  & AD                 & A                       & D                       & AD                       \\
\midrule
Baseline          &       & 0.99 & 0 & 0.02  & 0.89 & 0 & 0.02 & 0.95 & 0 & 0.02  \\\midrule
\multirow{4}{*}{FGSM} & $\epsilon=0.02$ & 0.9  & 0.07 & 0.06 & 0.04 & 0.04 & 0.06 & 0.6  & 0.01 & 0.08  \\
& $\epsilon=0.04$ & 0.64 & 0.27 & 0.21 & 0.02 & 0.06 & 0.09 & 0.35 & 0.16 & 0.18  \\
% & $\epsilon=0.06$ & 0.32 & 0.52 & 0.45 & 0.01 & 0.12 & 0.17 & 0.14 & 0.68 & 0.74 & 0.02 & 0.01 & 0    \\
& $\epsilon=0.08$ & 0.12 & 0.73 & 0.68 & 0.01 & 0.3  & 0.3  & 0.08 & 0.91 & 1     \\
& $\epsilon=0.1$  & 0.05 & 0.88 & 0.84 & 0.01 & 0.64 & 0.72 & 0.06 & 0.92 & 1     \\
% & $\epsilon=0.2$        & 0              & 0.99           & 0.53           & 0                  & 0.81               & 0.68               & 0.06                    & 0.91                    & 1                        & 0                        & 0.74                     & 0.7                      \\
% & $\epsilon=0.4$        & 0              & 0.99           & 0.1            & 0                  & 0.81               & 0.53               & 0.05                    & 0.92                    & 1                        & 0                        & 0.75                     & 0.2                      \\
% & $\epsilon=0.6$        & 0              & 0.99           & 0.07           & 0                  & 0.81               & 0.53               & 0.03                    & 0.91                    & 1                        & 0                        & 0.75                     & 0.1                      \\
% & $\epsilon=0.8$        & 0              & 0.99           & 0.07           & 0.01               & 0.81               & 0.65               & 0.02                    & 0.92                    & 1                        & 0                        & 0.75                     & 0.1                      \\
% & $\epsilon=1.0$        & 0              & 0.99           & 0.05           & 0.01               & 0.82               & 0.75               & 0.03                    & 0.92                    & 1                        & 0                        & 0.75                     & 0.1                      \\ \midrule
\midrule
\multirow{4}{*}{PGD} & $\epsilon=0.07$  & 0    & 0.63 & 0    & 0    & 0.79 & 0    & 0.01 & 0.94 & 0.97     \\
& $\epsilon=0.5$   & 0    & 0.64 & 0    & 0    & 0.79 & 0    & 0.01 & 0.94 & 0.9      \\
& $\epsilon=1.0$   & 0    & 0.64 & 0    & 0    & 0.79 & 0    & 0.01 & 0.94 & 0.9     \\
% & binary search   & 0    & 0 & 0    & 0    & 0 & 0    & 0 & 0 & 0     \\
\midrule
\multirow{3}{*}{DeepFool} & $i=1$ & 0.87           & 0.07           & 0.11           & 0.45               & 0.03               & 0.03               & 0.71                    & 0.02                    & 0.15                     \\
& $i=3$ & 0.1            & 0.13           & 0.21           & 0.02               & 0.03               & 0.04               & 0.4                     & 0.03                    & 0.46                     \\
& $i=5$ & 0              & 0.14           & 0.05           & 0                  & 0.03               & 0                  & 0.26                    & 0.05                    & 0.71                     \\
% & $i=7$ & 0              & 0.14           & 0              & 0                  & 0.03               & 0                  & 0.2                     & 0.07                    & 0.82                     & 0                        & 0.01                     & 0                        \\
% & $i=9$ & 0              & 0.14           & 0              & 0                  & 0.03               & 0                  & 0.16                    & 0.08                    & 0.89                     & 0                        & 0.01                     & 0           \\
\bottomrule
\end{tabular}
}
\caption{
    The max $n$-th percentile detector evaluation of LeNet5 (MNIST),
    M-CifarNet (CIFAR10), ResNet18 (CIFAR10).
    A means the accuracy on generated adversarial samples,
    D means the detection rate on adversarial samples that have fooled the models and
    AD represents the detection rates on adversarial samples that are correctly classified by the models.
    Detection ratios on clean evaluation data (false positives)
    in all cases are controlled less than $1\%$.}
\label{tab:defence_results}
\end{table}

\section{Evaluation}
\label{sec:evaluation}
\subsection{Attacks, Networks and Datasets}
We consider the Fast Gradient Sign Method (FGSM) \cite{goodfellow2014explaining},
Projected Gradient Descent~\cite{kurakin2016adversarial} and
Deepfool ~\cite{Moosavi15} attacks to evaluate the Taboo Trap.
We used the implementations of the attacks above from
Foolbox \cite{rauber2017foolbox} and implemented
our defense mechanism in Pytorch \cite{paszke2017automatic}.
At the time of writing,
FGSM is generally considered to be a weak adversary,
PGD a slightly stronger one and DeepFool the strongest attack.

We use LeNet5 \cite{lecun2015lenet} on MNIST \cite{lecun2010mnist}.
The LeNet5 model has 431K parameters and classifies MNIST hand-written
digits with an accuracy of $99.17\%$.
For the CIFAR10 \cite{krizhevsky2014cifar} datasets, we use
M-CifarNet \cite{zhao2018mayo} and
ResNet18 \cite{he2016deep} to perform classifications.
The M-CifarNet classifier \cite{zhao2018mayo} has 1.3M parameters and achieves $89.48\%$ classification accuracy, and
our ResNet18 model achieves $93.83\%$ accuracy on CIFAR10.

\subsection{Max \textit{n}th Percentile Detector}

Table \ref{tab:defence_results} shows the performance of the Taboo Trap
against three different attackers.
All of the networks have
shown a good detection rate (D and AD) for FGSMs with a relatively high $\epsilon$.
Interestingly, LeNet5 on MNIST and ResNet18 on CIFAR10 have shown
good accuracy for small values of $\epsilon$,
whereas M-CifarNet have not.
Both models have in common that they try to solve a
rather complex task with a much smaller architecture capability.
% This suggests that finding reported in \Fref{sec:reg_eff} can be made even stronger given a more capable architecture.
The Taboo Trap shows a relatively weak performance on a
relatively strong attack (PGD) and
a lot worse performance against a strong attack (DeepFool).
Similar to FGSM, the performance of the detector was better in models
with higher capacity for a given dataset.

To summarize, the $n$th percentile detector has shown relatively good detection rates against weak attackers
with the rates decreasing as the attacks become stronger.
Finally, the detector's performance and adversarial recovery effect
observed seems to be connected with the general capacity of the chosen DNN to solve a given task.

FGSM with a relatively large epsilon value can be seen as a
simple attack that is easy to execute at scale in practice.
It is a single-step method and a large epsilon makes the
adversarial samples transferable through different networks.
Iterative methods, such as DeepFool and PGD with binary search, can break our defense.
However, these attacks are usually performed with a strong assumption that attackers can iteratively optimize the adversarial samples, so detecting these expensive attacks is not the aim of our detection method.
As we have demonstrated in our results, we can efficiently
reject adversarial samples generated by cheap attacks with high detection ratios,
especially on adversarial samples that have successfully
fooled our models (the D measurement in \Fref{tab:defence_results}).

\subsection{Efficiency Comparison}

As mentioned previously, we designed Taboo Trap with efficiency in mind,
aiming at rejecting cheap attacks at scale.
\Fref{tab:len-act} shows the computational performance and also
defense efficiency when compared to MagNet \cite{Meng:2017:MTD:3133956.3134057}
and SafetyNet \cite{lu2017safetynet}.
Since Taboo Trap with maximum percentile as a transfer function is only
a value comparison in computation, it does not either introduce extra
compute or extra parameters.
Existing adversarial detection mechanisms have surprisingly large
overheads, some of them even have the defense networks to be
multiple times larger than the original classification network.
In the case of edge devices, when compute resources are limited and
the running networks are small, a large increase in compute and memory footprint
caused by defense is infeasible.
We realize Taboo Trap performs worse
on DeepFool compared to other methods.
However, we used a simple transfer function,
a more advanced transfer function might aid detection ratios.
Also, we highlight the purpose of designing Taboo Trap is to defeat
simple attacks at scale and Deepfool as an iterative attacking method
should not be considered as a cheap attack.
Taboo Trap detects simple attack like single-step FGSM
as well as other advanced detection methods (\Fref{tab:len-act}).

\begin{table}[!h]
\centering
\adjustbox{max width=\textwidth}{%
\begin{tabular}{@{}lllllll@{}}
\toprule
Technique& $\theta$ & FGSM & DeepFool & New MACs & as \% & New Params \\ \midrule
% Taboo Trap & $f=1^{st}$ max percentile & 0.99 & 0.14 & 0 & 0 & 0\\ \midrule
% MagNet & & 1.0 & 1.0 & 137592 & 6 & 297 \\ \midrule
% \multirow{2}{*}{Metzen et al.}& 1 layer, DeepFool & 0.92  & 0.82  & $\sim 34810489$ & 1518 & 5945367 \\
% & 3 layers, DeepFool & 0.98  & 0.92  & $\sim 21767179$ & 949 & 4861597 \\
Taboo Trap & $P_{1}$ & 0.99 & 0.14 & \textbf{0} & \textbf{0} & \textbf{0}\\ \midrule
MagNet \cite{Meng:2017:MTD:3133956.3134057} & & 1.0 & 1.0 & 190K & 20 & 297 \\ \midrule
\multirow{2}{*}{SafetyNet \cite{lu2017safetynet}}& 1 layer & 0.92  & 0.82  & 34.81M & 3622 & 5,95M \\
& 3 layers & 0.98  & 0.92  & 21.77M & 2265 & 4.86M \\
\bottomrule
\end{tabular}}
\caption{LeNet5 has 961K multiply-accumulate operations (MACs) and 17.75K parameters without defenses. SafetyNet is trained on the DeepFool adversarial samples. $P_{i}$ refers to the $i$th-percentile. FGSM $\epsilon=0.8$ is used. Overhead for MagNet is only calculated over the detector.}
\label{tab:len-act}
\end{table}

\section{Discussion}

% \begin{figure}[!h]
%     \centering
%     \includegraphics[width=\linewidth]{images/fmap.pdf}
%     \caption{Visualization of over-excited pixels with Taboo Trap. M-CifarNet is presented with FGSM ($\epsilon=0.8$) attack.}
%     \label{fig:fmap}
% \end{figure}

% \begin{figure*}[!ht]
%     \centering
%     \includegraphics[width=\linewidth]{images/transfer_functions.png}
%     \caption{
%     Different transform functions for activations.
%     $f_1$ is the maximum $1$st percentile, where the threshold $\alpha$ is different for each layer since is per-layer profiling dependent.
%     $f_2$ is restricted to $([0:1])$ and applied only on the first layer.
%     $f_2$ is restricted to $([0:1],[2:3],[4:5])$ and applied on all layers.
%     }
%     \label{fig:transform_func}
% \end{figure*}

\label{sec:discussion}
%We do not believe that the techniques presented in this work will accelerate the Arms Race.

%What makes our work particularly important is that we have presented a simple, low-cost way to defend networks against GreyBox attacks, and a combination of transforms and behaviour-level constraints may extend this protection to BlackBox/WhiteBox attacks.

In 1883, the cryptographer Auguste Kerckhoffs enunciated a design principle that
has stood the test of time: a system should withstand enemy capture,
and in particular it should remain secure if everything about it,
except the value of a key, becomes public knowledge~\cite{Kerckhoffs1883}.
Here, we have studied the performance of using the $n$th maximum percentile as a simple transform function. This can be applied to any randomly chosen subset of the activation values, and gives plenty of opportunity to add the equivalent of a cryptographic key.

In the grey-box setup, attackers might be aware of the deployment of the taboo trap classifier but remain ignorant of the chosen transform function.
In a white-box scenario, attackers might construct samples that trick the defence employed -- they can try to feed in various inputs to figure out the transform function.
The defence against white-box attacks is changing the key --
for example, to combine a series of transform functions that work on different activation values or different ranges. That way, a vendor can produce two different versions of his product, one for the mass market and one for industrial use; an attacker who develops adversarial samples for the former will not be able to use them on the latter.

\begin{table}[!h]
\centering
\adjustbox{max width=0.7\linewidth}{
\begin{tabular}{@{}lllll@{}}
\toprule
\textbf{Attack}  & ~~~~~~~~~$\theta$   & \textbf{$f_1$} & \textbf{$f_2$} & \textbf{$f_3$} \\ \midrule
FGSM & $\epsilon=0.4$ & 0.94 & 0.97  &   \textbf{0.98} \\
PGD & $\epsilon=0.07, i=5$ & \textbf{0.94} & 0.92  & 0.61 \\
DeepFool & $i=5$ & 0.01 & \textbf{0.12} & 0.06 \\ \bottomrule
\end{tabular}
}
\caption{Taboo Trap on LeNet5 with three different transform functions: $f_1$ and $f_3$ on all layers; $f_2$ on the first.}
\label{tab:lenet_different_tf}
\end{table}

So how do we choose good keys?
We have experimented with
a number of different transform functions and
have observed them exhibiting a variety of defensive capabilities.

% In \Fref{fig:fmap}, we show a heatmap on the over-threshold pixels
% on different adversarial samples.
% We apply the attack on the evaluation dataset to generate adversarial samples
% and record which pixels are over-driven.
% The brighter the pixel is, the more often this happened.
% On a particular transform function (max-percentile),
% some channels show a large number of sensitive pixels
% but some channels only show one or two.
% Those channels and layers can be hard to identify
% unless the defense has been deployed
% and tested against real adversarial samples.

To demonstrate that
behavioral defense can work with different transform functions,
Table \ref{tab:lenet_different_tf} shows the
defensive capabilities of LeNet5 (MNIST)
trained with three different transform functions,
$f_1$ is the maximum $1$st percentile ($[0:\alpha]$),
where the threshold $\alpha$ is different for each layer since it is per-layer profiling dependent.
$f_2$ is restricted to $(0:1]$ and applied only on the first layer.
$f_2$ is restricted to $[0:1],[2:3],[4:5]$ and applied on all layers.
All of the networks were trained to have an accuracy
of around $~99\%$ with a false positive rate of less than $1\%$.
As can be seen in the table, the performance of different
attacks on various keys differs from each other.
% the performance of DeepFool against $f_1$ is six times
% that of $f_3$ and ten times that of $f_2$.
% Yet despite bad performance with $f_1$,
% it has the best detector performance against PGD.
% It is only marginally better than $f_2$, but is $1.5$x better than $f_3$.
% Poor performance against PGD does not stop $f_3$ from scoring best against FGSM.
This suggests that each of the attacks focuses
on exploiting a particular layer within a particular range,
and so a defender can choose different families of
transform functions to block different adversaries.

The diversity and heterogeneity of the potential defensive transform functions gives the classic solution to the white-box attack.
They make defense unpredictable, and the defender can respond to both the expectation and the experience of attacks by changing a key.
The key selects from among different transform functions and will also add non-determinism to grey-box attacks, where the transform functions are known but the parameters are not.

% Detection can also be randomised.
% In our evaluation we used the canonical binary detector --
% the sample is either adversarial or not.
% One can make use a continuous function instead,
% measure how confident we are
% (e.g. how far away it is from the threshold in case of $n$th percentile function)
% that a particular sample is adversarial, and respond in an appropriate but non-deterministic way.

We designed Taboo Trap for
deployment on low-cost hardware where computation may be severely limited,
for example by battery life.
%, in such cases the implementation needs a certain amount of care.
% One way to save energy is use the detector per-layer and abort computation when the detector fires.
% However, this might introduce a timing side-channel where the implementation detail makes this relevant.
However, nothing stops the defender from using classifiers with Taboo Traps in conjunction with other strategies where the computation budget permits this.
As our approach does not change the network structure or add any other additional components to the pipeline
it is easily combined with other, more expensive, defensive systems like MagNet \cite{DBLP:journals/corr/abs-1711-08478} or SafetyNet \cite{lu2017safetynet}.

\section{Conclusion}

In this paper we presented the Taboo Trap,
a radically new way to detect adversarial samples that is both simple and cheap.
% Instead of learning a network's behaviour,
% we restrict it during training to avoid certain taboo values,
% whether taboo outputs or taboo activations of intermediate layers,
% and use taboo violation to detect adversarial inputs.
% We explored the trade-offs between the detectability and aggressiveness
% of the resulting restrictions,
% and found that our models show consistently
% dependable behaviour on a set of common computer-vision benchmarks.
% We explained how to train the instrumented networks
% and showed that the classifiers are still able to provide
% comparable performance to the baseline models regardless of how complex the models are.
The run-time compute overhead of Taboo Trap is close to zero,
with the only additional cost being a slight increase in the time
taken to train the network.
We evaluated our simple mechanism and showed that it
performs well against a range of popular attacks.
The simple $n$th percentile transfer function that we tested
did not perform as well as more complex detection mechanisms
such as MagNet but still provides a very useful tool in the defender's armoury.

In addition to simplicity and low cost,
the Taboo Trap offers diversity in defence.
It can be used with a wide variety of transfer functions,
which perform much the same function that key material does
in cryptographic systems:
the security of a system need not reside in keeping the basic design secret,
but in the secrecy of parameters that can be chosen at random for each implementation.
This opens the prospect of extending defences against adversarial machine-learning attacks
from black-box applications to grey-box and even white-box adversaries.

\section*{Acknowledgements}
\textit{Partially supported with funds from Bosch-Forschungsstiftungim Stifterverband}

\bibliographystyle{abbrv}
\bibliography{egbib}

\end{document}